# Applications of Large Scale Foundation Models for Autonomous Diving

Yu Huang, Yue Chen, and Zhu Li, *Senior Member, IEEE*

*Abstract*—Since DARPA's Grand Challenges (rural) in 2004/05 and Urban Challenges in 2007, autonomous driving has been the most active field of AI applications. Recently powered by large language models (LLMs), chat systems, such as chatGPT and PaLM, emerge and rapidly become a promising direction to achieve artificial general intelligence (AGI) in natural language processing (NLP). There comes a natural thinking that we could employ these abilities to reformulate autonomous driving. By combining LLM with foundation models, it is possible to utilize the human knowledge, commonsense and reasoning to rebuild autonomous driving systems from the current long-tailed AI dilemma. In this paper, we investigate the techniques of foundation models and LLMs applied for autonomous driving, categorized as simulation, world model, data annotation and planning or E2E solutions etc.

## I. INTRODUCTION

Autonomous Driving has been active for more than 10 years. In 2004 and 2005, DARPA held the Grand Challenges in rural driving of driverless vehicles. In 2007, DAPRA also held the Urban Challenges for autonomous driving in street environments. Then professor S. Thrun at Stanford university, the first-place winner in 2005 and the second-place winner in 2007, joined Google and built Google X and the self-driving team.

Autonomous driving, as one of the most challenging applications of AI with machine learning and computer vision etc., actually has been shown to be a "long tailed" problem, i.e. the corner cases or safety-critical scenarios occur scarcely[1-3].

Foundation models [33] have taken shape most strongly in NLP. On a technical level, foundation models are enabled by transfer learning and scale. The idea of transfer learning is to take the "knowledge" learned from one task and apply it to another task. Foundation models usually follow such a paradigm that a model is pre-trained on a surrogate task and then adapted to the downstream task of interest via fine-tuning, shown in Fig. 1.

Most of the Large Scale Language Models (LLMs) [52] appearing recently are among or based on the Foundation Models. Recent models with billion parameters, like GPT-3/4 [23, 49], have been effectively utilized for zero/few-shot learning, achieving impressive performance without requiring large-scale task-specific data or model parameter updating.

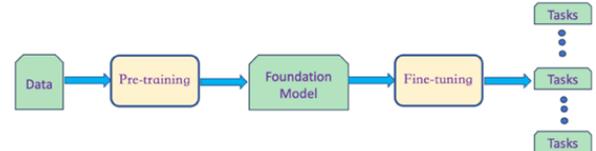

Fig. 1. Foundation Model

The pretraining tasks can be divided into five categories according to the learning methods: Mask Language Modeling (MLM), Denoising AutoEncoder (DAE), Replaced Token Detection (RTD), Next Sentence Prediction (NSP), Sentence Order Prediction (SOP).

LangChain is a framework designed to simplify the creation of applications using LLMs[4]. As a language model integration framework, LangChain's use-cases largely overlap with those of language models in general, including document analysis and summarization, chatbots, and code analysis.

The overwhelming success of Diffusion Models[133] starts from image synthesis but extends to other modalities, like video, audio, text, graph and 3-D model etc. As a new branch of multi-view reconstruction, NeRF (Neural Radiance Field) [183,198] provides implicit representation of 3D information. Marriage of diffusion models and NeRF has achieved remarkable results in text-to-3D synthesis.

There are two survey papers[253, 255] for large language model-based autonomous driving (as well with intelligent transportation in [253]). However, we try to investigate this area from a new point of view, also a broader domain.

In this paper, we first review the LLMs and their extension to visual language models, multi-modal LLMs and embodied agents, as well as two related techniques, i.e. NeRF and diffusion models. Then we investigate the applications of foundation models and LLMs to autonomous driving from the backend and the frontend. The backend utilization includes simulation and annotation, and the frontend appliance consists of world models and planning/decision making, as well as E2E driving operations.

## II. LARGE SCALE LANGUAGE MODEL

The vanilla Transformer [14] is firstly proposed with an encoder-decoder architecture, designed for extracting

This work was supported in part by the Futurewei Technology Inc.
Yu Huang is with the Roboraction.AI, Portland, Oregon 97227 USA (e-mail: yu.huang07@gmail.com).

Yue Chen with Futurewei Technology Inc, Santa Clara, CA95050 USA(yue.chen@futurewei.com). Zhu Li is with the Department of Computer Science & Electrical Engineering, University of Missouri, Kansas city, MO 64110 USA (e-mail: zhu.li@ieee.org).

information from natural language. The basic building block is called a cell, which is composed of two modules, Multi-head Attention (MHA), and a feed-forward network (FFN), shown in Fig. 2.

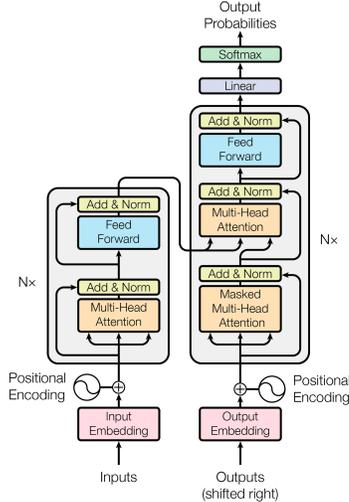

Fig. 2. Transformer model architecture [14]

The MHA is a module that runs multiple independent self-attention layers in parallel to capture advanced semantics of inputs across various feature levels. This enables jointly attending to information from different representational subspaces and across different parts of the sequence. The FFN is a feature extractor that projects the advanced semantics from different MHA modules to the same feature space.

The transformer model does not use recurrence or convolution and treats each data point as independent of the other. Hence, positional information is added to the model explicitly to retain the information regarding the order of words in a sentence. Positional encoding (PE) is the scheme through which the knowledge of the order of objects in a sequence is maintained.

There are some modifications of Transformer architecture to improve its efficiency and scalability, like MQA [20], Switch Transformers[25], RoPE [29], FlashAttention1/2 [40, 58] in Megatron-LM (a large, powerful transformer developed by NVIDIA) [6] and lightLMM (a Python-based LLM inference and serving framework with lightweight design, easy scalability, and high-speed performance) [13], GQA [55], and PageAttention [67] in vLLM (a high-throughput and memory-efficient inference and serving engine for LLMs)[12] etc.

Hugging Face is a company that focuses on NLP and provides a variety of tools and resources for working with NLP models [5]. One of their notable contributions is the development of the Transformers library, which is an open-source library that provides pre-trained models and various tools for working with SoTA NLP models, including those based on machine learning and deep learning techniques.

A variant called multi-query attention (MQA) is proposed in [20], where the keys and values are shared across all of the different attention "heads", greatly reducing the size of these tensors and hence the memory bandwidth requirements of incremental decoding. The resulting models can indeed be much faster to decode, and incur only minor quality degradation from the baseline. Grouped-query attention (GQA) [55] is a generalization of MQA which uses an intermediate (more than one, less than number of query heads) number of key-value heads.

Mixture of Experts (MoE) models select different parameters for each incoming example. The result is a sparsely-activated model with a number of parameters but a constant computational cost. Switch Transformer [25] applies the simplified MoE routing algorithm and builds intuitive improved models with reduced communication and computational costs.

A variant named Rotary Position Embedding(RoPE) is proposed in [29] to effectively leverage the positional information, shown in Fig. 3. The RoPE encodes the absolute position with a rotation matrix and meanwhile incorporates the explicit relative position dependency in self-attention formulation. Notably, RoPE enables valuable properties, including the flexibility of sequence length, decaying inter-token dependency with increasing relative distances, and the capability of equipping the linear self-attention with relative position encoding. The enhanced transformer with rotary position embedding is called RoFormer.

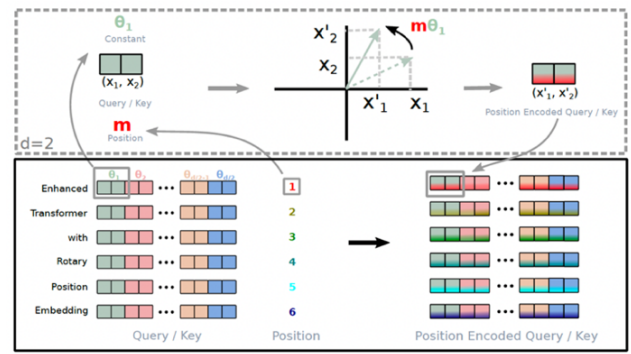

Fig. 3. Implementation of RoPE [29].

FlashAttention [40] is an IO-aware exact attention algorithm that uses tiling to reduce the number of memory reads/writes between GPU high bandwidth memory (HBM) and GPU on-chip SRAM. FlashAttention-2 [58], with better work partitioning to address these issues is proposed. PagedAttention is proposed by vLLM [67], being an attention algorithm inspired by the classical virtual memory and paging techniques in operating systems.

LLMs are the category of Transformer-based language models that are characterized by having an enormous number of parameters[17, 21, 26], typically numbering in the hundreds of billions or even more[23, 37, 43, 49]. These models are trained on massive text datasets, enabling them to understand natural language and perform a wide range of complex tasks, primarily through text generation and comprehension. Some well-known examples of LLMs include GPT-3/4[23, 49], PaLM[39], OPT[42], and LLaMA1/2[48, 59] (shown in Fig. 4).

Extensive research has shown that scaling can largely improve the model capacity of LLMs. Thus, it is useful to establish a quantitative approach to characterizing the scaling effect. There are two representative scaling laws for Transformer language models: one from OpenAI [22], another from Google DeepMind[38].

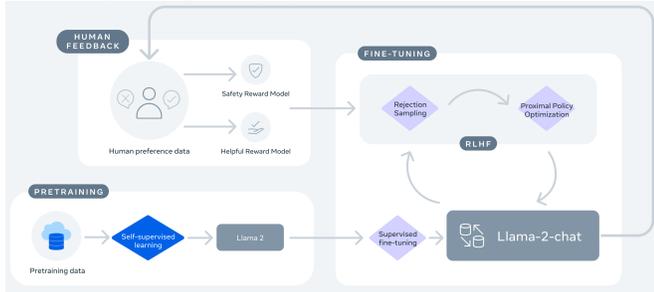

Fig. 4. LLaMA 2 Training [59]

There is a power-law relationship [22] between model performance and each of the following three factors: the number of non-embedding model parameters $N$, the training dataset size in tokens $D$, and the amount of non-embedding compute $C$. There exists an optimal budget allocation between the model size and the amount of data processed in training. They demonstrate that within a fixed compute budget (the term "compute budget" refers to the total amount of computations), the optimal model performance is obtained by training very large models and stopping early before convergence.

Another scaling law [38] claims that given the compute budget, the number of data tokens processed in training should be scaled equally to the size of the model. They show that smaller models that are adequately trained can overperform undertrained large models. The above work summarizes the empirical laws for deciding the size of the dataset under a fixed budget.

Data parallelism [47] distribute the whole training corpus into multiple GPUs with replicated model parameters and states. Data parallel running techniques can be split into two categories: asynchronous and synchronous data parallelism.

All-gather and all-reduce communication patterns are often used in data parallelism. All-gather patterns let every processor communicates its data to every other processor. All-reduce patterns are a layer on top of all-gather combining aggregation with summing or averaging.

The well-known asynchronous method is Parameter Server, where one server saves a baseline set of parameters while distributed workers keep model replicas that train on different mini-batches. The popular synchronous data parallelism method is Distributed Data Parallelism (DDP). DDP clones a model and allocates copies to $m$ different workers. An initial "global minibatch" is used, then split evenly across the replicas to make local gradient updates. These gradients are then aggregated across replicas to generate an entire update, typically using an all-reduce communication pattern.

Model parallelism [47] is the technique of splitting, or sharding, a neural architecture graph into subgraphs, and each subgraph, or model shard, is assigned to an individual GPU. These shards might correspond to groups of stacked layers in a feedforward network. Its speedup rate relies highly on the architecture and the sharding strategy.

Tensor parallelism [53] is a technique used to assign a big model into a number of GPUs. When the input tensors are multiplied with the first weight tensor, matrix multiplication is the same to the weight tensor column-wise splitting, individual multiplication of each column with the input, and then concatenation of the split outputs. It then transfers these outputs from the GPUs and concatenates them together to get the final result. Most tensor parallel operators require at least one all-gather communication step to reaggregate partitioned outputs.

In Pipeline parallelism [53] the incoming batches are partitioned into mini-batches, and the layers of the model are split across multiple GPUs, thus a pipeline is created at each stage, i.e. the results of the previous stage is taken as input by a set of contiguous layers and passed downstream, which allows different GPUs in parallel to take part in the computational process. This has realized the lowest communications and can be arranged across nodes.

Zero Redundancy Optimizer (ZeRO) [17] is currently an important technique for training large-scale models. ZeRO optimizes redundant model states (i.e. optimizer states, gradients, and parameters) in memory by partitioning them in three corresponding stages across processors and optimizing the communication, with the final model states evenly split on each node. On the second stage (ZeRO-2) which partitions both the optimizer states and the gradients, ZeRO-Offload [26] is built, which offloads the data and computations of both states to the CPU, thus leveraging the CPU to save the GPU memory.

However, due to limited GPU memory, the number of parameters stored in GPU memory and copied over all devices is still limited, and sub-optimal data partitioning and limited PCIe bandwidth require large batch training for efficiency. Further, Rajbhandari et al. [28] show ZeRO-Infinity, a heterogeneous system technology that leverages CPU and NVMe memory (which is cheap, slow, but massive) in parallel across multiple devices to aggregate efficient bandwidth for current GPU clusters, shown in Fig. 5.

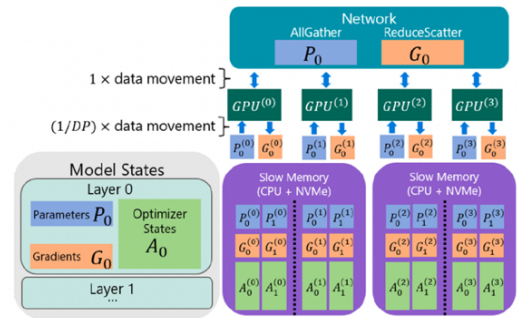

Fig. 5. ZeRO-Infinity [28]

The current large-scale deep learning models often reach 10B or even 100B parameters, such as GPT3 [23] (175 B). Training such large models usually require a comprehensive use of data parallelism, tensor parallelism, pipeline parallelism, mixed precision training, and ZeRO-like distributed optimizers, also known as 3D hybrid parallelism. Usually, tensor parallelism with the greatest communication volume is prioritized within a single node. Data parallelism is also placed within a node if possible to speed up the gradient communication, or mapped to a nearby node.

Megatron-LM [6, 19] is a large transformer lib developed by NVIDIA, with a model parallelism framework in training. TensorRT-LLM [7] is a Python API to define LLMs and builds TensorRT engines that contain state-of-the-art optimizations to perform inference efficiently on NVIDIA GPUs, transitioned from FasterTransformer[8].

DeepSpeed [9-10] is a well-known open-source library for large model training in PyTorch, which supports ZeRO, 3D-parallelism, etc. Colossal-AI [35] leverages a series of parallel methods to generate distributed AI models by training. Besides of the mentioned 1D tensor parallelism, Colossal-AI also combines 2D, 2.5D, 3D tensor parallelism strategies and sequence parallelism (long sequence modeling by breaking the memory wall along the large sequence dimension).

The "pre-train+fine-tune" procedure is replaced by another procedure called "pre-train+prompt+predict". In this paradigm, instead of adapting pre-trained LMs to downstream tasks via objective engineering, downstream tasks are reformulated to look more like those solved during the original LM training with the help of a textual prompt.

In this way, by selecting the appropriate prompts, the model behavior can be manipulated so that the pre-trained LM itself can be used to predict the desired output, sometimes even without any additional task-specific training. Prompt engineering [45] works by finding the most appropriate prompt to allow a LM to solve the task at hand.

The emergent abilities of LLMs are one of the most significant characteristics that distinguish them from smaller language models. Specifically, in-context learning (ICL)[46], instruction following [60] and reasoning with chain-of-thought (CoT) [66] are three typical emergent abilities for LLMs.

ICL employs a structured natural language prompt that contains task descriptions and possibly a few task examples as demonstrations. Through these task examples, LLMs can grasp and perform new tasks without necessitating explicit gradient updates. Instruction-tuning and following aims to teach models to follow natural language (including prompt, positive or negative examples, and constraints etc.), to perform better multi-task learning on training tasks and generalization on unseen tasks. CoT takes a different approach by incorporating intermediate reasoning steps, which can lead to the final output, into the prompts instead of using simple input-output pairs.

Parameter-efficient fine tuning (PEFT) [31, 50, 68] is a crucial technique used to adapt pre-trained language models (LLMs) to specialized downstream applications. PEFT can be divided into addition-based, selection/specification-based or reparameterization-based. Adapters [18] add domain specific layers between neural network modules. They propose to add fully-connected networks after attention and FFN layers in Transformer. Unlike the transformer FFN block, Adapters usually have a smaller hidden dimension than the input.

Li & Liang [24] develop the idea of *soft prompts* with a distinctive flavor, called prefix-tuning. Instead of adding a soft prompt to the model input, trainable parameters are prepended to the hidden states of all layers. Another method, P-tuning v1 [27] leverages few continuous free parameters to serve as prompts fed as the input to the pre-trained language models.

Then the continuous prompts are optimized using gradient descent as an alternative to discrete prompt searching.

An empirical finding is that properly optimized prompt tuning can be comparable to fine-tuning universally across various model scales and NLU tasks. The improved method P-tuning v2 given in [34] can be viewed as an optimized and adapted implementation, designed for generation and knowledge probing, shown in Fig. 6. Liu et al. [41] propose a parameter-efficient method, called (IA)$^3$, to learn three vectors which rescale key, value, and hidden FFN activations respectively. Training only these three vectors for each transformer block leads to high parameter efficiency.

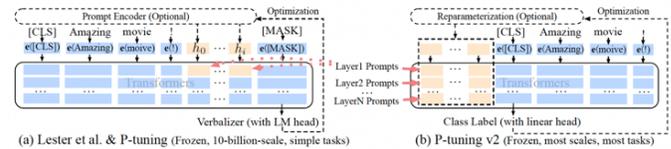

Fig. 6. Comparison of P-tuning v1 and P-tuning v2 [34]

Lester et al. [30] explore prompt tuning, a method for conditioning language models with learned soft prompts, which achieves competitive performance compared to full fine-tuning and enables model reuse for many tasks. Hu et al. [32] proposed utilizing low-rank decomposition matrices (LoRA) to reduce the number of trainable parameters needed for fine-tuning language models. Modifications of LoRA occur like AdaLoRA[51].

Note: In the field of model compression for LLM research [61], researchers often combine multiple techniques with low-rank factorization, including pruning, quantization and so on. As research in this area continues, there may be further developments in applying low-rank factorization to compressing LLMs, like QLoRA [56] (shown in Fig. 7), but there is still ongoing exploration and experimentation required to fully harness its potential for LLMs.

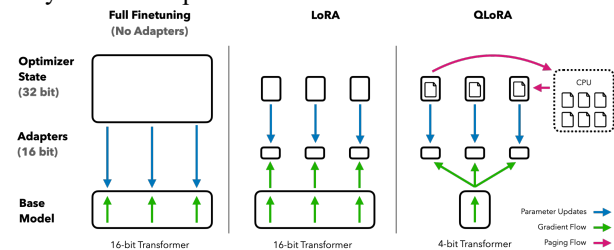

Fig. 7. Full Fine-tuning vs. LoRA vs. QLoRA [56]

Since LLMs are trained to capture the data characteristics of pre-training corpora (including both high-quality and low-quality data), they are likely to generate toxic, biased, or even harmful content for humans. It is necessary to align LLMs with human values, e.g., helpful, honest, and harmless.

Reinforcement Learning from Human Feedback (RLHF) [65] has emerged as a key strategy for fine-tuning LLM systems to align more closely with human preferences. Ouyang et al. [37] introduce a human-in-the-loop process to create a model that better follows instructions, shown in Fig. 8. Bai et al. [44] propose a method for training a harmless AI assistant without human labels, providing better control over AI behavior with minimal human input.

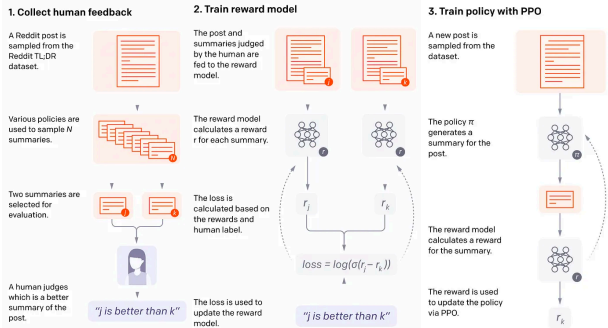

Fig. 8. RLHF in GPT-3.5 [37]

Hallucination is a big problem of LLMs to avoid, which refers to a situation where the model generates content that is not based on factual or accurate information [64]. Hallucination can occur when the model produces output that includes details, facts, or claims that are fictional, misleading, or entirely fabricated, rather than providing reliable and truthful information.

Hallucination can be unintentional and may result from various factors, including biases in the training data, the model's lack of access to real-time or up-to-date information, or the inherent limitations of the model in comprehending and generating contextually accurate responses.

Explainability [63] refers to the ability to explain or present the behavior of models in human-understandable terms. Improving the explainability of LLMs is crucial. With that, end users are able to understand the capabilities, limitations, and potential flaws of LLMs. Besides, explainability acts as a debugging aid to quickly advance model performance on downstream tasks.

From the application view, LLMs can handle high-level reasoning tasks such as question answering and commonsense reasoning. Understanding exclusive abilities of LLMs in in-context learning and chain-of-thought prompting, as well as the phenomenon of hallucination, are indispensable to explaining and improving models.

Evaluation is of paramount prominence to the success of LLMs [57]. Evaluating LLMs helps to better understand the strengths and weakness. Additionally, better evaluations can provide a better guidance for human-LLMs interaction, which could inspire future interaction design and implementation.

Moreover, the broad applicability of LLMs underscores the paramount importance of ensuring their safety and reliability, particularly in safety-sensitive sectors such as financial institutions and healthcare facilities. Finally, as LLMs are becoming larger with more emergent abilities, existing evaluation protocols may not be enough to evaluate their capabilities and potential risks.

Recently, RAG (retrieval-augmented generation) [36] has gained popularity in NLP due to the rise of general-purpose LLMs[36]. RAG typically consists of two phases: retrieving contextually relevant information, and guiding the generation process using the retrieved knowledge.

RAG methods offer a promising solution for LLMs to effectively interact with the external world. With the help of external knowledge, LLMs can generate more accurate and reliable responses. The most common method is to use a search engine as a retriever such as New Bing. Due to the vast amount of information available on the Internet, using a search engine can provide more real-time information.

Knowledge Graph (KG) [62] is a semantic network comprising entities, concepts, and relations, which can catalyze applications across various scenarios such as recommendation systems, search engines, and question-answering systems. Some works use LLMs to augment KGs for, e.g., knowledge extraction, KG construction, and refinement, while others use KGs to augment LLMs for, e.g., training and prompt learning, or knowledge augmentation.

## III. VISUAL LANGUAGE MODEL, MULTI-MODAL LLM AND EMBODIED AGENT

The Transformer architecture has also made significant contributions to the computer vision community. For instance, it has inspired the development of models like Vision Transformer (ViT) [70] shown in Fig. 9, and its extensions [90, 97].

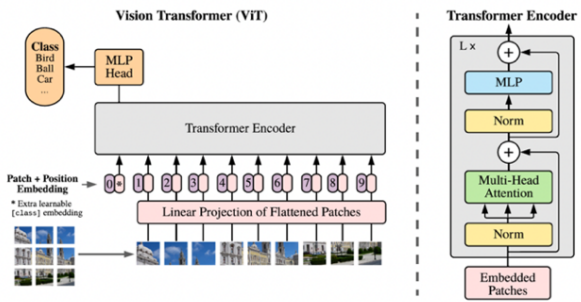

Fig. 9. Vision Transformer (ViT) [70]

Pix2Seq [76] casts object detection as a language modeling task conditioned on the observed pixel inputs. Object descriptions (e.g., bounding boxes and class labels) are expressed as sequences of discrete tokens, and a neural network is trained to perceive the image and generate the desired sequence.

SAM (segment anything model) [95] is a foundation model for image promptable segmentation, consists of an image encoder, a flexible prompt encoder, and a fast mask decoder. SEEM [96] is a promptable, interactive model for Segmenting Everything Everywhere all at once in an image. The visual foundation model SEAL [106] is capable of segmentation any point cloud sequences, shown in Fig. 10.

Vision-Language Models (VLMs) bridge the capabilities of Natural Language Processing (NLP) and Computer Vision (CV), breaking down the boundaries between text and visual information to connect multimodal data, such as CLIP(Contrastive Language–Image Pre-training) [71] (shown in Fig. 11), BLIP-1/2[74, 87], Flamingo[77] and PaLI-1/X /3[81, 105, 121] etc.

BLIP [74] is a Visual Language Pre-training framework which transfers flexibly to both vision-language understanding and generation tasks. BLIP-2 [87] bootstraps vision-language pre-training from off-the-shelf frozen pre-trained image encoders and frozen large language models. The All-Seeing model (ASM) [109] is a unified location-aware image-text foundation model. The model consists of two key components: a location-aware image tokenizer and an LLM-based decoder.

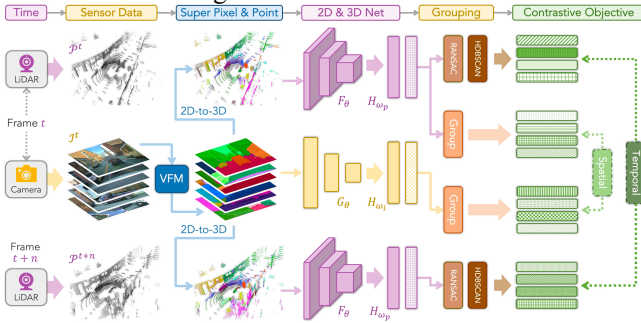

Fig. 10. SEAL [106]

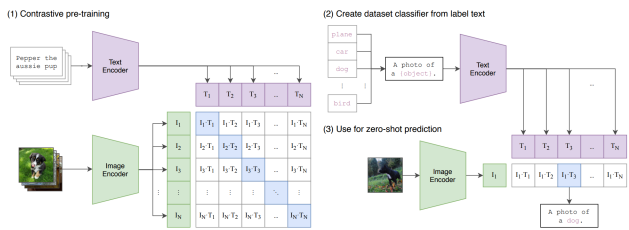

Fig. 11. CLIP model [71]

Motivated by the potential of LLMs, numerous multimodal LLMs (MLLMs) [92, 99, 116, 117] have been proposed to expand the LLMs to the multimodal field, i.e., perceiving image/video input, and conversating with users in multiple rounds. Pre-trained on massive image/video-text pairs, the above models can only handle image-level tasks, such as image captioning and question answering.

Building on the powerful pretrained LLM weights, multimodal LLMs aim to handle multiple types of input beyond text. Multimodal LLMs have been widely applied to various tasks, such as image understanding, video understanding, medical diagnosis, and embodied AI etc.

The main architectural idea of PaLM-E [92] is to inject continuous, embodied observations such as images, state estimates, or other sensor modalities into the language embedding space of a pre-trained language model. This is realized by encoding the continuous observations into a sequence of vectors with the same dimension as the embedding space of the language tokens, shown in Fig. 12.

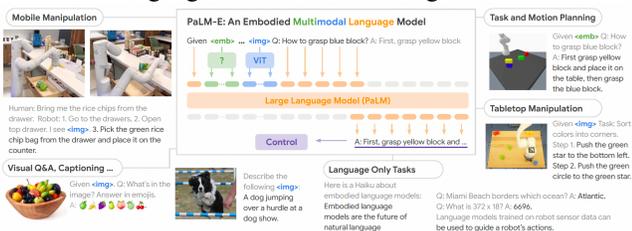

Fig. 12. PaLM-E [92]

Note: VLMs and MLLMs are also able to be fine-tuned like LLMs, such as Visual Prompt [75], LLaVA (instruction tuning) [94] and InstructBLIP [100].

Leveraging multimodal information for 3D modality could be promising to improve 3D understanding under the restricted data regime. PointCLIP [79] conducts alignment between CLIP encoded point cloud and 3D category texts. Specifically, a point cloud is encoded by projecting it into multi-view depth maps without rendering, and aggregate the view-wise zero shot prediction to achieve knowledge transfer from 2D to 3D. PointCLIP V2 [83] is a 3D open-world learner, to fully unleash the potential of CLIP on 3D point cloud data, in which large-scale language models are leveraged to automatically design a more descriptive 3D-semantic prompt for CLIP's textual encoder.

ULIP [84] is proposed to learn a unified representation of images, texts, and 3D point clouds by pre-training with object triplets from these three modalities. It leverages a pre-trained vision-language model and then learns a 3D representation space aligned with the common image-text space, using a small number of automatically synthesized triplets. ULIP v2 [101] is a tri-modal pre-training framework that leverages state-of-the-art large multimodal models to automatically generate holistic language counterparts for 3D objects, shown in Fig. 13. It does not require any 3D annotations, and is therefore scalable to large datasets.

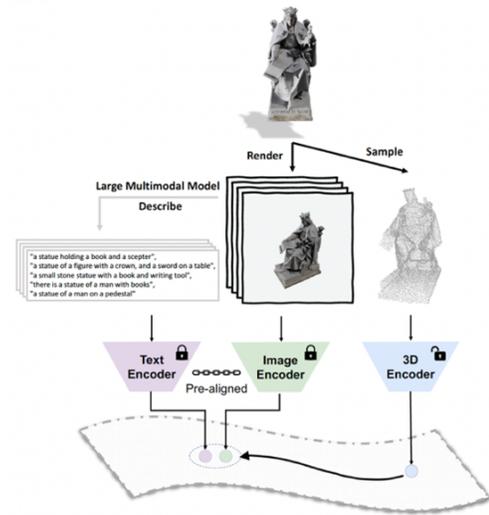

Fig. 13. ULIP v2 [101]

CLIP2Scene [86] transfers CLIP knowledge from 2D image-text pre-trained models to a 3D point cloud network. A Semantic-driven Cross-modal Contrastive Learning framework is designed, that pre-trains a 3D network via semantic and spatial-temporal consistency regularization. OpenShape [102] is a method for learning multi-modal joint representations of text, image, and point clouds by multi-modal contrastive learning.

World models have been on research for a long history in control engineering and artificial intelligence. Explicitly the knowledge of an agent about its environment is represented in the world model, in which a defined generative model is tailored to predict the next observation given past observations and the current action [111, 217, 228-229, 240, 250, 258-261]. The main use cases are: representation learning, planning, or learning a policy (neural simulator). Dynalang [111] is an agent that learns a multimodal world model to predict future text and image representations and learns to act from imagined model rollouts, which learning is shown in Figure 14.

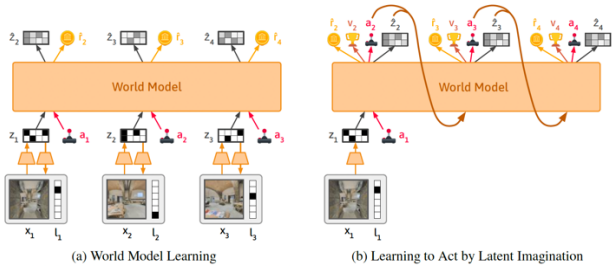
Fig. 14. World Model [111]

World modeling can be regarded as a pre-training task for supervised learning of a concise and generic representation, as a state for conventional reinforcement learning (RL) methods, which accelerated training speed. Look-ahead search can employ world models to plan by predicting the future actions. Also, world models can act as an environment simulator to handle the RL sampling efficiency issues.

Since LLMs are still confined to token-level, left-to-right decision-making processes during inference, a framework for language model inference, Tree of Thoughts (ToT)[103], which generalizes over the popular CoT approach to prompting language models, and enables exploration over coherent units of text that serve as intermediate steps toward problem solving.

Graph of Thoughts (GoT)[110] is a framework that advances prompting capabilities in LLMs. The key idea and primary advantage of GoT is the ability to model the information generated by an LLM as an arbitrary graph, where units of information ("LLM thoughts") are vertices, and edges correspond to dependencies between these vertices.

COT escalates the number of query requests, leading to increased costs, memory, and computational overheads. Algorithm of Thoughts [112] is a strategy that propels LLMs through algorithmic reasoning pathways, pioneering a mode of in-context learning (ICL). By employing algorithmic examples, the innate recurrence dynamics of LLMs are exploited with merely one or a few queries.

Some techniques to employ external tools to compensate for the deficiencies of LLMs [91, 98, 108]. A general tool learning framework can be formulated as follows: starting from understanding the user instruction, models should learn to decompose a complex task into several subtasks, dynamically adjust their plan through reasoning, and effectively conquer each sub-task by selecting appropriate tools.

The success of LLMs is undoubtedly exciting as it demonstrates the extent to which machines can learn human knowledge. In advanced tasks with LLMs, the translation of natural language input into actionable results is crucial, such as GATO [78], ReAct [82] and RT-1/2/X[85, 107, 122].

The guiding design principle of GATO is to train on the widest variety of relevant data possible, including diverse modalities such as images, text, proprioception, joint torques, button presses, and other discrete and continuous observations and actions, shown in Fig. 15. To enable processing this multi-modal data, they serialize all data into a flat sequence of tokens. In this representation, GATO can be trained and sampled from akin to a standard large-scale language model. During deployment, sampled tokens are assembled into dialogue responses, captions, button presses, or other actions based on the context.

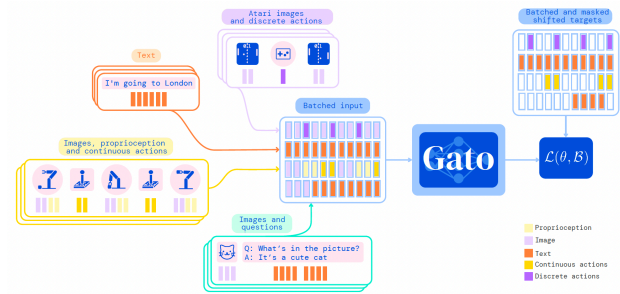
Fig. 15. GATO (A Generalist Agent) [78]

ReAct [82] explores the use of LLMs to generate both reasoning traces and task-specific actions in an interleaved manner to a diverse set of language and decision making tasks. ReAct is used in [224] to perceive and analyze its surrounding environment for autonomous driving. Robotics Transformer 1 (RT-1) [85] can absorb large amounts of data, effectively generalize, and output actions at real-time rates for practical robotic control. It takes a short sequence of images and a natural language instruction as input and outputs an action for the robot at each time step. RT-2 [107] is a family of models derived from fine-tuning large vision-language models trained on web-scale data to directly act as generalizable and semantically aware robotic policies, shown in Fig. 16. RT-X [122] is a high-capacity model, trained on a large scale robotics data, and exhibits positive transfer and improves the capabilities of multiple robots by leveraging experience from other platforms.

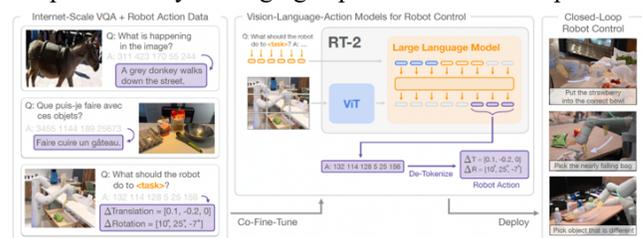
Fig. 16. RT-2 (Vision-Language-Action Models) [107]

LLM based agents [113, 115] can exhibit reasoning and planning abilities comparable to symbolic agents through techniques like Chain-of-Thought (CoT) and problem decomposition. They can also acquire interactive capabilities with the environment, akin to reactive agents, by learning from feedback and performing new actions.

Recent works have developed more efficient reinforcement learning agents for robotics and embodied AI [88, 104]. In Embodied AI/agents, AI algorithms and agents no longer learn from datasets, instead learn through interactions with environment from an egocentric perception. The focus is on enhancing agents' abilities for planning, reasoning, and collaboration in embodied environments. Some approaches combine complementary strengths into unified systems for embodied reasoning and task planning. High-level commands enable improved planning while low-level controllers translate commands into actions. Dialogue for information gathering can accelerate training. Some agents can work for embodied decision-making and exploration guided by internal world models.

Habitat [69, 73, 123] is a simulation platform for training virtual robots in interactive 3D environments and complex physics-enabled scenarios. Based on that, comprehensive contributions go to all levels of the embodied AI stack - data, simulation, and benchmark tasks.

LM-Nav [80] is a system for robotic navigation, constructed entirely out of pre-trained models for navigation (ViNG), image-language association (CLIP), and language modeling (GPT-3), without requiring any fine-tuning or language-annotated robot data.

"Describe, Explain, Plan and Select" (DEPS) [89] is an interactive planning approach based on LLMs. It helps with better error correction from the feedback during the long-haul planning, while also bringing the sense of proximity via goal Selector, a learnable module that ranks parallel sub-goals based on the estimated steps of completion and improves the original plan accordingly.

Any-Modality Augmented Language Model (AnyMAL) [114] is a unified model that reasons over diverse input modality signals (i.e. text, image, video, audio, IMU motion sensor), and generates textual responses. AnyMAL inherits the powerful text-based reasoning abilities of LLMs, and converts modality-specific signals to the joint textual space through a pre-trained aligner module.

## IV. DIFFUSION MODEL

In recent years, the diffusion model has achieved great success in the community of image synthesis [133]. It aims to generate images from Gaussian noise via an iterative denoising process. Its implementation is built based on strict physical implications, which consists of a diffusion process and a reverse process. In the diffusion process, an image is converted to a Gaussian distribution by adding random Gaussian noise with iterations. The reverse process is to recover the image from the distribution by several denoising steps.

Diffusion models represent a family of probabilistic generative models that progressively introduce noise to data and subsequently learn to reverse this process for the purpose of generating samples, shown in Fig. 17. These models have recently garnered significant attention due to their exceptional performance in various applications, setting new benchmarks in image synthesis, video generation, and 3D content generation. The fundamental essence of diffusion-based generative models lies in their capacity to comprehend and understand the intricacies of the world.

There are three foundational diffusion models that are widely utilized, including DDPM [127], NCSNs [125] and SDE[126]. Among them, NCSNs (*Noise Conditional Score Network*)[125] seeks to model the data distribution by sampling from a sequence of decreasing noise scales with the annealed Langevin dynamics. In contrast, DDPM (Denoising Diffusion Probabilistic Models)[127] models the forward process with a fixed process of adding Gaussian noise, which simplifies the reverse process of the diffusion model into a solution process for the variational bound objective. These two basic diffusion models are actually special cases of score-based generative models [126]. SDE (Stochastic Differential Equation)[126], as the unified form, models the continuous diffusion and reverse with SDE. It proves that the NCSNs and DDPM are only two separate discretization styles of SDE.

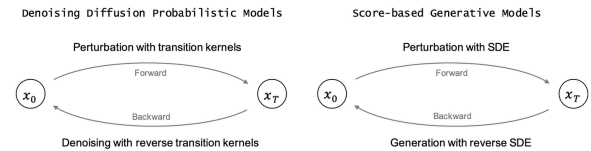

Fig. 17. Diffusion Model [133]

Diffusion models have also gained success in a wide range of other domains, including sequence generation, decision-making, planning and character animation. Experimental evidence also showed that training with synthetic data generated by diffusion models can improve task performance on tasks.

Latent diffusion models (LDM) [131] are a type of diffusion model that models the distribution of the latent space of images and have recently shown remarkable performance in image synthesis. The LDM consists of two models: an autoencoder and a diffusion model, shown in Fig. 18. The autoencoder learns to compress and reconstruct images using an encoder and a decoder. The encoder first projects the image to a lower dimensional latent space, and the decoder then reconstructs the original image from the latent space. Then the latent generative model is trained to recreate a fixed forward Markov chain via DDPMs [127].

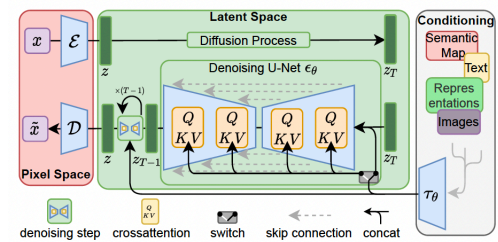

Fig. 18. Latent Diffusion [131] (Stable Diffusion[124])

Note: Like LLMs, there are some fine-tuning methods of diffusion models, such as ControlNet [140] and T2I Adapter [142].

GLIDE [130] introduces ablated diffusion model (ADM) into text-to-image generation. Compared to previous diffusion based methods, GLIDE uses larger model with 3.5B parameters and larger pairwise datasets, which achieved better results on many benchmarks. Different from GLIDE,

Imagen [134] combines a frozen T5 language model with a super-resolution diffusion model. The frozen encoder will encode the text instruction and generates an embedding, then the first diffusion model will accordingly generate an low-resolution image. The second diffusion model accepts this image with the text embedding and outputs a high-resolution image.

DALL-E-2 [132] is extension of DALL-E[72] and it combines CLIP encoder with diffusion decoder for image generation and editing tasks, shown in Fig. 19. Compared with Imagen, DALL-E-2 leverages a prior network to translation between text embedding and image embedding. DALL-E3[153] is an approach addressing prompt following: caption improvement. First it learns a robust image captioner which produces detailed, accurate descriptions of images. Then it applies this captioner to produce more detailed captions.

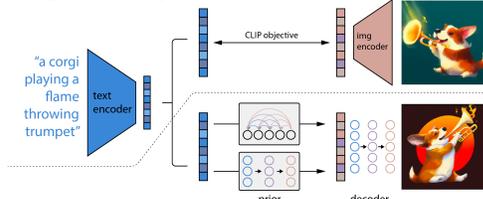

Fig. 19. DALL-E-2 [132]

A method called Point-E in [138] for 3D object generation is proposed. First it generates a single synthetic view using a text-to-image diffusion model, and then produces a 3D point cloud using a second diffusion model which conditions on the generated image. LidarCLIP [139] is proposed, a mapping from automotive point clouds to a pre-existing CLIP embedding space. Using image-lidar pairs, a point cloud encoder is supervised with the image CLIP embeddings, effectively relating text and lidar data with the image domain as an intermediary. 3DFuse[143] is a framework that incorporates 3D awareness into pretrained 2D diffusion models, enhancing the robustness and 3D consistency of score distillation-based methods. It first constructs a coarse 3D structure of a given text prompt and then utilizing projected, view-specific depth map as a condition for the diffusion model.

A method of Fantasia3D [145] for high-quality text-to-3D content creation is proposed, shown in Fig. 20. The key to Fantasia3D is the disentangled modeling and learning of geometry and appearance. For geometry learning, it relies on a hybrid scene representation, encoding surface normal extracted from the representation as the input of the image diffusion model. For appearance modeling, the spatially varying bidirectional reflectance distribution function (BRDF) is introduced into the text-to-3D task to learn the surface material for photorealistic rendering of the generated surface.

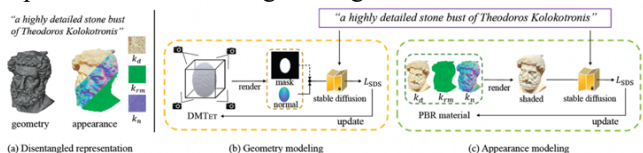

Fig. 20. Fantasia3D [145]

NExT-GPT[152] is an end-to-end general-purpose any-to-any MM-LLM system. Connecting an LLM with multimodal adaptors and different diffusion decoders, NExT-GPT is enabled to perceive inputs and generate outputs in arbitrary combinations of text, images, videos, and audio. Moreover, a modality-switching instruction tuning (MosIT) is introduced by which a high-quality dataset for MosIT is built by manually curating.

EasyGen[154] is a model designed to enhance multimodal understanding and generation by harnessing the capabilities of diffusion models and large language models (LLMs). EasyGen is built upon a bidirectional conditional diffusion model named BiDiffuser, which promotes more efficient interactions between modalities. EasyGen handles image-to-text generation by integrating BiDiffuser and an LLM via a simple projection layer.

V. NEURAL RADIANCE FIELD

NeRF (Neural Radiance Field) [156] enables photorealistic synthesis in a 3D-aware manner. Developed in 2020 by researchers at the University of California, Berkeley, NeRF uses deep neural networks to model the 3D geometry and appearance of objects in a scene, enabling the creation of high-quality visualizations that are difficult or impossible to achieve using traditional rendering techniques.

The key idea of NeRF is to encode the appearance of a scene as a function of 3D location and viewing direction, called the radiance field. The radiance field explains how light goes through the space and interacts with object surfaces. It can be leveraged to synthesize images from any chosen viewpoints.

The NeRF algorithm involves several steps: data acquisition, network training, and rendering. Fig. 21 gives an illustration of the NeRF scene representation and differentiable rendering pipeline. It displays the steps performed in synthesizing images, which consists of sampling 5D coordinates along camera rays, applying an MLP to estimate color and volume density, and aggregating these values into an image by volume rendering.

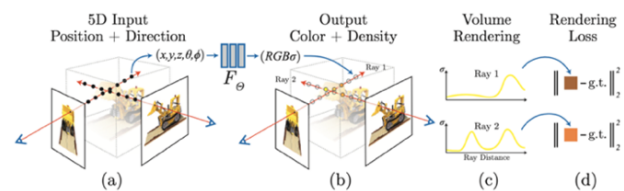

Fig. 21. The NeRF pipeline [156]

Different from traditional 3D reconstruction techniques that represent scenes using explicit expressions such as point clouds, grids, and voxels, NeRF samples each ray and retrieves the 3D location of each sampling point and the 2D viewing direction of the ray. Subsequently, these 5D vector values are fed into a neural network to determine the color and volume density of each sampling point. NeRF constructs a field parameterized by an MLP neural network to continuously optimize parameters and reconstruct the scene.

NeRF++ [157] gives a "inverted sphere" parameterization of space for NeRF to large-scale, unbounded 3D scenes. Points outside the unit sphere are inverted back and passed through a separate MLP. MVSNeRF [158] is a generic deep neural network to reconstruct radiance fields from only three nearby input views via network inference. It applies plane-swept cost volumes for geometry-aware scene inference, and it is combined with physically based volume rendering for NeRF reconstruction. NeRF in the Wild [160] considers additional modules to the MLP representation to handle inconsistent lighting and objects across different images.

Neural Scene Graphs (NSG) [161] is a novel view synthesis method from monocular videos captured while driving (ego-vehicle views). It decomposes a dynamic scene with multiple moving objects into a learned scene graph that includes individual object transformations and radiances. Thus, each object and the background are encoded by individual neural networks. Further, the sampling of the static node is limited to layered planes for efficiency, i.e., a 2.5D representation.

PixelNeRF [162] predicts a continuous neural scene representation conditioned on one or few input images from learning, where a MLP produces color and density fields for NeRF to render. When they are trained on multiple scenes, scene priors are learned for reconstruction, then high fidelity reconstruction of scenes from a few views is enabled. Similarly, features are extracted from several context views in Stereo Radiance Fields [159], where learned correspondence matching between pairwise features across context images are leveraged to aggregate features across context images. Finally, IBRNet [163] introduces transformer networks across the ray samples to reason about visibility.

Several different methods have been proposed for speeding up volumetric rendering of MLP-based representations. KiloNeRF [166] combines empty space skipping and early termination with a dense 3D grid of MLPs, each with a much smaller number of weights than a standard NeRF network.

A scene is spatially decomposed and dedicate smaller networks are built for each decomposed part in DeRF [164]. When they work together, the whole scene can be rendered. Regardless of the number of decomposed parts, the enables near-constant inference time. The output depths from NeRF are directly supervised in Depth-supervised NeRF [165] (in the form of depth along each ray) using the sparse point cloud output, as a byproduct of camera pose estimation with the structure-from-motion (SfM) technique. A pretrained sparse-to-dense depth completion network is applied directly in [179] to sparse SfM depth estimates, then depth is used to both guide sample placement and supervise the depth produced by NeRF, shown in Fig. 22.

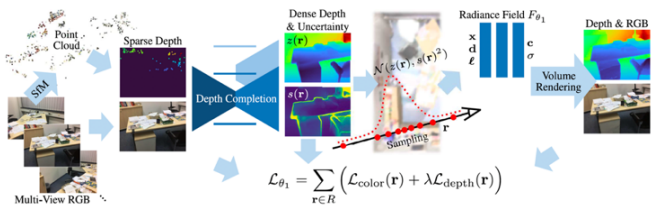

Fig. 22. Radiance field optimization [179]

The volume density is tied to an signed distance field in NeuS [170] and the transmittance function is re-parameterized to achieve its maximal slope precisely at the zero-crossing of this SDF, which allows an unbiased estimate of the corresponding surface.

A neural scene rendering system[168], called Object NeRF, learns an object compositional neural radiance field and produces realistic rendering with editing capability for a clustered and real world scene. NeRF is extended in [169] to jointly encode semantics with appearance and geometry, named Semantic NeRF, so that complete and accurate 2D semantic labels can be achieved using a small amount of in-place annotations specific to the scene. A Single View NeRF (SinNeRF) framework [174] consists of thoughtfully designed semantic and geometry regularizations. Panoptic Neural Fields (PNF) [178] is an object-aware neural scene representation that decomposes a scene into a set of objects (things) and background (stuff).

Mip-NeRF [167] adjusts the positional encoding of 3D points to take in account the pixel footprint, see Fig. 23. It pre-integrates the positional encoding over a conical frustum corresponding to each quadrature segment sampled along the ray. Mip-NeRF 360 [175] extends MipNeRF and addresses issues that arise when training on unbounded scenes. It applies a non-linear scene parametrization, online distillation, and a distortion based regularizer. StreetSurf [199] extends prior object centric neural surface reconstruction techniques to the unbounded street views that are captured with non-object-centric, long and narrow camera trajectories.

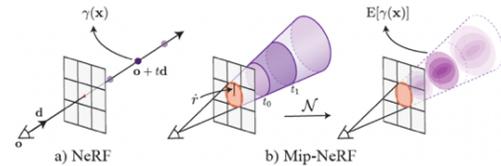

Fig. 23. NeRF vs Mip-NeRF [167]

Recently, there has been tremendous progress in driving scene simulation using NeRF. Block-NeRF [176] achieves city-scale reconstruction by modeling the blocks of cities with several isolated NeRFs to increase capacity, shown in Fig. 24. While BlockNeRF uses a fixed grid of blocks, Mega-NeRF [173] uses a dynamic grid that is adapted to the scene being rendered. This makes Mega-NeRF a more scalable and efficient framework from large-scale visual captures using NeRF.

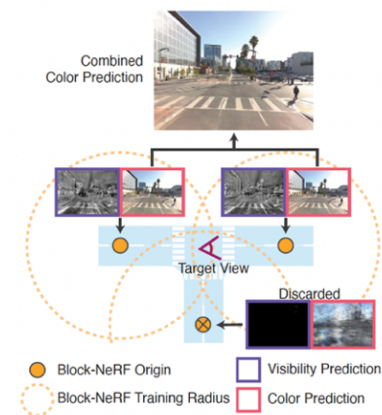

Fig. 24. Block-NeRF [176]

URF [177] exploits additional LiDAR data to supervise the depth prediction. In these large-scale scenarios, special care must be taken to handle the sky and the highly varying exposure and illumination changes. Several large scale scene rendering methods, i.e. S-NeRF[191], SUDS [201], MatrixCity [206] and UE4-NeRF [209] are proposed.

$D^2$NeRF [184] is an approach for generating high quality NeRF models of static scenes. $D^2$NeRF learns a 3D scene representation using separate NeRFs for moving objects and the static background. FEGR [202] learns to intrinsically decompose the driving scene for applications such as relighting. Lift3D [203] use NeRF to generate new objects and augment them to driving datasets, demonstrating the capability of NeRF to improve downstream task performance. Lift3D is applied in [230] to generate physically realizable adversarial examples of driving scenarios.

Dream Fields [172] generates 3D models from natural language prompts, while avoiding the use of any 3D training data, shown in Fig. 25. Specifically, Dream Fields optimizes a NeRF from many camera views such that rendered images score highly with a target caption according to a pre-trained CLIP model.

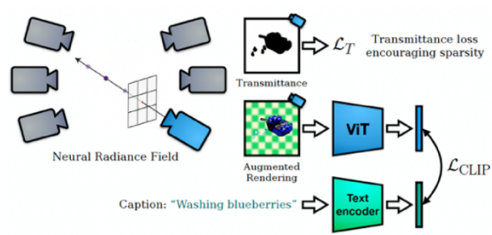

Fig. 25. Dream Fields [172]

Text2NeRF[195] generates a wide range of 3D scenes with complicated geometric structures and high-fidelity textures purely from a text prompt. NeRF is adopted as the 3D representation and a pre-trained text-to-image diffusion model is leveraged to constrain the 3D reconstruction of the NeRF to reflect the scene description, where a monocular depth estimation method is employed to offer the geometric prior. This method requires no additional training data but only a natural language description of the scene as the input.

3D synthesis by the diffusion model requires large-scale datasets of labeled 3D data and efficient architectures for denoising 3D data. DreamFusion [180] circumvents these limitations by using a pretrained 2D text-to-image diffusion model, shown in Fig. 26. A loss based on probability density distillation is used in a DeepDream-like procedure, it optimizes a randomly-initialized 3D model (NeRF) via gradient descent such that its 2D renderings from random angles achieve a low loss.

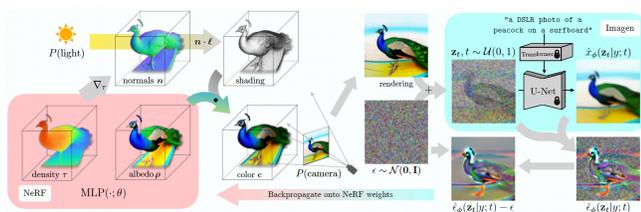

Fig. 26. DreamFusion [180]

To assist and direct the 3D generation, Latent-NeRF [185] operates directly in the latent space of the Latent Diffusion Model. This allows further refinement in RGB space, where shading constraints is introduced or further guidance from RGB diffusion models is applied. NeRDi [187] is a single-view NeRF synthesis framework with general image priors from 2D diffusion models. Formulating single-view reconstruction as an image-conditioned 3D generation problem, the NeRF representations are optimized by minimizing a diffusion loss on its arbitrary view renderings with a pretrained image diffusion model under the input-view constraint.

Magic-3D[186] creates high quality 3D mesh models by utilizing a two-stage optimization framework, shown in Fig. 27. First, a coarse model is obtained using a low-resolution diffusion prior and accelerated with a sparse 3D hash grid structure. Using the coarse representation as the initialization, a textured 3D mesh model with an efficient differentiable renderer interacting with a high-resolution latent diffusion model, is optimized.

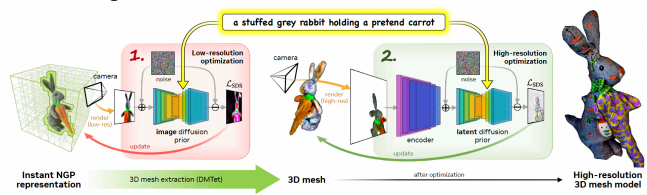

Fig. 27. Magic-3D [186]

Ponder [189] is a method to self-supervised learning of point cloud representations by differentiable neural rendering. It trains a point cloud encoder within a devised point-based neural renderer by comparing the rendered images with real images on massive RGB-D data. Ponder v2 [208] is a 3D pre-training framework designed to facilitate the acquisition of efficient 3D representations. It proposes an universal paradigm to learn point cloud representations, serving as a bridge between 3D and 2D worlds. It trains a point cloud encoder within a devised volumetric neural renderer by comparing the rendered images with the real images.

For the problem of reconstructing a full 360 photographic model of an object from a single image, RealFusion [190] takes an off-the-self conditional image generator based on diffusion and engineer a prompt that encourages it to "dream up" novel views of the object. Using DreamFusion [180], it fuses the given input view, the conditional prior, and other regularizers in a final, consistent reconstruction. Make-It-3D [192] employs a two-stage optimization pipeline: the first stage optimizes a neural radiance field; the second stage transforms the coarse model into textured point clouds and further elevates the realism with diffusion prior.

Shap-E [194] is a conditional generative model for 3D assets, which directly generates the parameters of implicit functions that can be rendered as both textured meshes and neural radiance fields, trained with a conditional diffusion model. HiFA [197] unlocks the potential of the diffusion prior. To improve 3D geometry representation, auxiliary depth supervision is applied for NeRF-rendered images and the density field of NeRFs is regularized. Points-to-3D [205] is a framework to bridge the gap between sparse yet freely available 3D points and realistic shape-controllable 3D generation, shown in Fig. 28. It consists of three parts: a scene representation model (NeRF), a 2D diffusion model (ControlNet[140]), and a point cloud 3D diffusion model (Point-E[138]).

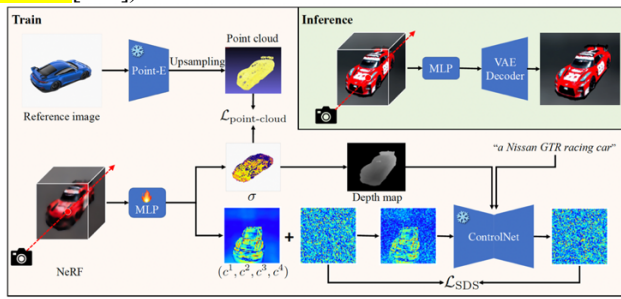

Fig. 28. Points-to-3D [205]

## VI. APPLICATIONS OF LLMS AND FOUNDATION MODELS FOR AUTONOMOUS DRIVING

Autonomous driving develops techniques for vehicles to navigate without human manipulating, aiming at reducing accidents to ennhance traffic efficiency. The Society of Automotive Engineers (SAE) defines the driving automation level from Level 0 (No Automation) to Level 5 (Full Automation). Currently, most commercial vehicles are at Level 2 or 3, and for some special ODD (operation domain design), L4 is realized, such as robotaxi.

We can categorize the exiting autonomous driving solutions broadly into the modular paradigm and the end-to-end system, illustrated in Fig. 29. So far, there still exist some serious challenges like robustness, interpretability, generalization, and safety/security etc. Building a safe, stable, and explainable AD system is still an open problem.

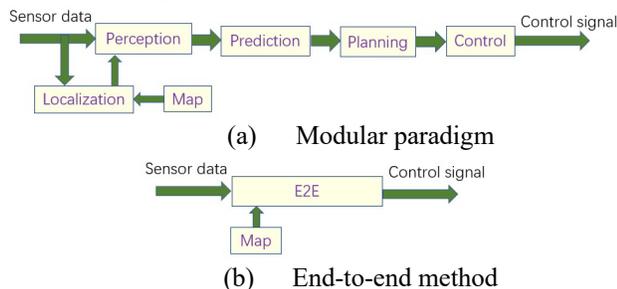

(a) Modular paradigm

(b) End-to-end method

Fig. 29. Autonomous driving solutions

Perception collects information from sensors and discovers relevant knowledge from the environment. It develops a contextual understanding of driving environment, such as detection, tracking and segmentation of obstacles, road signs/marking and free space drivable areas. Based on the sensors implemented, the environment perception task can be tackled by using LIDARs, cameras, radars or a fusion between these three kinds of devices.

As the core of autonomous driving systems, the decision making module is a crucial module. The output of the decision module can be low level control signals, such as the throttle, speed, acceleration, etc., or it can be high-level signals, such as the action primitive and planned trajectory for the planning module.

Motion planning is a core challenge in autonomous driving, aiming to plan a driving trajectory that is safe and comfortable. The intricacies of motion planning arise from its need to accommodate diverse driving scenarios and make reasonable driving decisions.

Existing motion planning approaches generally are classified as two categories. The rule-based method makes explicit rules to estimate driving trajectories. The kind of methods is clearly interpretable but generally works hard to process scarce driving scenarios not explained easily by rules. The alternative method is learning-based, belonging to a data-driven style, which learns models from large-scale human driver's navigating trajectories.

While they obtain good performance, the interpretability is sacrificed due to a blackbox modeling scheme. Basically, both popoular rule-based and learning-based solutions lack of the common sense reasoning ability of human drivers, which restricts their capabilities in handling the long-tailed problem of autonomous driving.

Below we categorize the foundation models' application in autonomous driving based on its grounding levels, from simulation (data synthesis), world model (learning-and-then-prediction), perception data annotation (auto-labeling), and decision making or driving actions (E2E). In the simulation area, we split it further into two directions: sensor data synthesis and traffic flow generation. In the decision making or driving actions, the approaches are classified as three groups: LLMs' integration, tokenization like GPT and pre-trained foundation models, shown in Fig. 30.

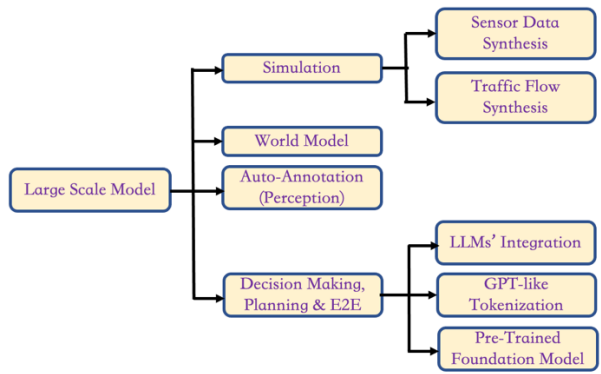

Fig. 30. Categories of AV foundation models

We include methods with diffusion models and NeRFs applied in autonomous driving. Though they may not apply LLMs or foundation models yet now, but the potential in nature and forseeable bindings make us believe its coming true in future, with the techniques from Dream fields[172], Dreamfusion[180], Latent-NeRF[185], Magic-3D[186], NeRDi[187], Text2NeRF[195], NExT-GPT[152], DALL-E3[152] and EasyGEN[154] etc.

arXiv 2311.12144, Nov. 20, 2023## A. Simulation and World Model

We group simulation and world model together since world models could be regarded as a neural simulator. Simulation is kind of methods for AIGC (AI generated content), but focusing on static and dynamic components in the driving environment. World models understand the dynamics and then predict the future.

### 1. Sensor Data Synthesis

READ (Autonomous Driving scene Render)[216] is a large-scale neural rendering method to synthesize the autonomous driving scene. In order to represent driving scenarios, an $\omega$-net rendering network is proposed to learn neural descriptors from sparse point clouds. This model can not only synthesize realistic driving scenes but also stitch and edit driving scenes.

Scene-Diffusion[219] is a learned method of traffic scene generation designed to simulate the output of the perception system of a self-driving car. Inspired by latent diffusion, a combination of diffusion and object detection is used to directly create realistic and physically plausible arrangements of discrete bounding boxes for agents.

MARS (ModulAr and Realistic Simulator)[225] is a modular framework for photorealistic autonomous driving simulation based on NeRFs. This open-sourced framework consists of a background node and multiple foreground nodes, enabling the modeling of complex dynamic scenes.

UniSim[227] is a neural sensor simulator that takes a single recorded log captured by a sensor-equipped vehicle and converts it into a realistic closed-loop multi-sensor simulation. UniSim builds neural feature grids to reconstruct both the static background and dynamic actors in the scene, and composites them together to simulate LiDAR and camera data at new viewpoints, with actors added or removed and at new placements, shown in Fig. 31. To better handle extrapolated views, it incorporates learnable priors for dynamic objects, and leverages a convolutional network, called hypernet, to complete unseen regions.

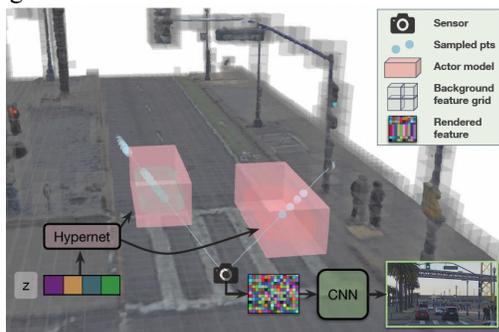

Fig. 31. UniSim [227]

Adv3D [230] is proposed for modeling adversarial examples as NeRFs. It is trained by minimizing the surrounding objects' confidence predicted by 3D detectors on the training set. To generate physically realizable adversarial examples initialized from Lift3D[203], primitive-aware sampling and semantic-guided regularization that enable 3D patch attacks with camouflage adversarial texture are proposed.

DriveSceneGen [239] is a data-driven driving scenario generation method that learns from the real world driving dataset and generates entire dynamic driving scenarios from scratch. The pipeline consists of two stages: a generation stage and a simulation stage. In the generation stage, a diffusion model is employed to generate a rasterized Birds-Eye-View (BEV) representation. In the simulation stage, the vectorized representation of the scenario is consumed by a simulation network based on the Motion TRansformer (MTR) framework.

MagicDrive[241] is a street view generation framework offering diverse 3D geometry controls, shown in Fig. 32, including camera poses, road maps, and 3D bounding boxes, together with textual descriptions, achieved through tailored encoding strategies. The power of pre-trained stable diffusion is harnessed and further fine-tuned for street view generation with road map information by ControlNet [140].

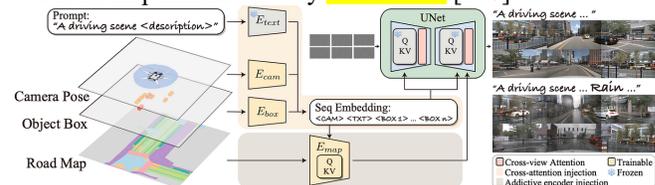

Fig. 32. MagicDrive [241]

DrivingDiffusion [247] is a spatial-temporal consistent diffusion framework, to generate realistic multi-view videos controlled by 3D layout. It is based on the widely used image synthesis diffusion model where the 3D layout is utilized as additional control information (this is also a drawback). Based on CLIP, local prompt to guide the relationship between the whole image and local instances, and global prompt, are cooperated. Unfortunately, it is not yet a E2E simulation used for autonomous driving.

### 2. Traffic Flow Synthesis

Realistic Interactive TrAffic flow (RITA) [212] is an integrated component of existing driving simulators to provide high-quality traffic flow. RITA consists of two modules called RITABackend and RITAKit. RITABackend is built to support vehicle-wise control and provide diffusion-based traffic generation models with from real-world datasets, while RITAKit is developed with easy-to-use interfaces for controllable traffic generation via RITABackend.

CTG (controllable traffic generation) [221] is a conditional diffusion model for users to control desired properties of trajectories at test time (e.g., reach a goal or follow a speed limit) while maintaining realism and physical feasibility through enforced dynamics. The key technical idea is to leverage diffusion modeling and differentiable logic to guide generated trajectories to meet rules defined using signal temporal logic (STL). It can be extended as guidance to multi-agent settings and enable interaction-based rules like collision avoidance.

CTG++[222] is a scene-level conditional diffusion model guided by language instructions, shown in Fig. 33. A scene-level diffusion model equipped with a spatio-temporal transformer backbone is designed, which generates realistic and controllable traffic. Then a LLM is harnessed to convert a user's query into a loss function, guiding the diffusion model towards query-compliant generation.

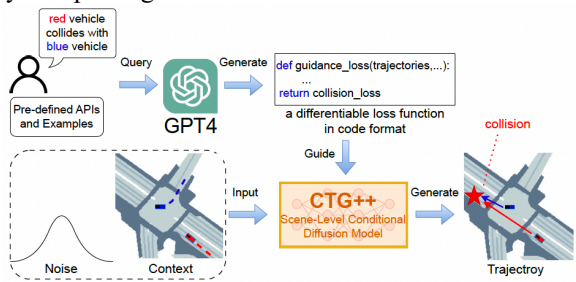

Fig. 33. CTG++ model [222]

SurrealDriver [232] is a generative 'driver agent' simulation framework based on LLMs, capable of generating human-like driving behaviors: understanding situations, reasoning, and taking actions. Interviews with 24 drivers are conducted to get their detailed descriptions of driving behavior as CoT prompts to develop a 'coach agent' module, which can evaluate and assist 'driver agents' in accumulating driving experience and developing humanlike driving styles.

*3. World Model*

World models hold great promise for generating diverse and realistic driving videos, encompassing even long-tail scenarios, which can be utilized to train foundation models in autonomous driving. Furthermore, the predictive capabilities in world models facilitate end-to-end driving, ushering in a new era of seamless and comprehensive autonomous driving experiences.

Anomaly Detection is an important issue for data closed loop of autonomous driving, which decides the data selection efficiency for model upgrade by training with newly selected valuable data. An overview of how world models can be leveraged to perform anomaly detection in the domain of autonomous driving is given in [226].

TrafficBots[217] is a multi-agent policy built upon motion prediction and end-to-end driving. Based on that, a world model is obtained and tailored for the planning module of autonomous vehicles. To generate configurable behaviors, for each agent both a destination as navigational information and a time-invariant latent personality to specify the behavioral style are introduced. To improve the scalability, a scheme of positional encoding for angles is designed, allowing all agents to share the same vectorized context based on dot-product attention. As a result, all traffic participants in dense urban scenarios are simulated.

UniWorld[229], a spatial-temporal world model, is able to perceive its surroundings and predict the future behavior of other participants. UniWorld involves initially predicting 4D geometric occupancy as the World Models for foundational stage and subsequently fine-tuning on downstream tasks. UniWorld can estimate missing information concerning the world state and predict plausible future states of the world.

GAIA-1 ('Generative AI for Autonomy') [240], , shown in Fig. 34, proposed by Wayve (an UK startup), is a generative world model that leverages video, text, and action inputs to build realistic driving scenarios while a fine-grained control over ego-vehicle behavior and scene features is given. The world modeling is casted as an unsupervised sequence modeling problem, where the inputs are mapped to discrete tokens, and the next token is predicted in the sequence.

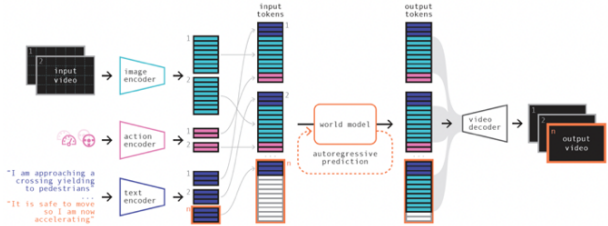

Fig. 34. GAIA-1 ('Generative AI for Autonomy') [240]

DriveDreamer[250] is a world model derived from real-world driving scenarios. It is seen that modeling the world in driving scenes needs a huge search space, the diffusion model is tailored to generate a representation of the environment, which is named as Auto-DM, where the noise is estimated in the diffusion steps to get loss for optimizing the model.

In [256] a world modeling approach is proposed by an AD startup Waabi, which first tokenizes sensor observations with VQVAE and then predicts the future via discrete diffusion, shown in Fig. 35. To efficiently decode and denoise tokens in parallel, Masked Generative Image Transformer （MaskGIT） is recast into the discrete diffusion framework with a few simple changes.

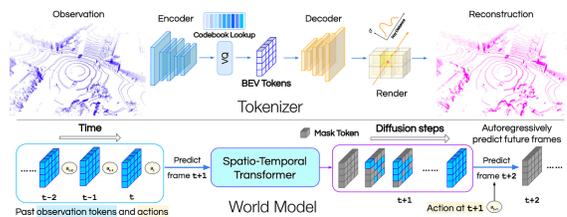

Fig. 35. VQVAE + diffusion [256]

MUVO [258] is a Multimodal World Model with Geometric VOxel Representations to take into account the physical attributes of the world. It utilizes raw camera and lidar data to learn a sensor-agnostic geometric representation of the world, to predict raw camera and lidar data as well as 3D occupancy representations multiple steps into the future, conditioned on actions.

Based on the vision-action pairs, a general world model based on MLLM and diffusion model for autonomous driving, termed ADriver-I, is constructed in [259]. It takes the vision-action pairs as inputs and autoregressively predicts the control signal of current frame. The generated control signals together with the historical vision-action pairs are further conditioned to predict the future frames. With the predicted next frame, ADriver-I performs further control signal prediction.

OccWorld is a world model explored in [260]. It works in the 3D Occupancy space to predict the ego vehicle movement and how the scenes evolve. They propose a reconstruction-based scene tokenizer on the 3D occupancy to get scene tokens for the surrounding scenes. They a GPT-like spatial-temporal generative transformer is adopted to build scene and ego car tokens for the future occupancy and ego trajectory.

A driving world model, Drive-WM, is proposed in [261] for multi-view video generation and end-to-end planning. Particularly it enables inferring future based on driving maneuvers and estimating the trajectory.

*4. Discussions*

Table I summarizes the methods of simulation and world model given in session VI.A, including modalities, functions and technologies applied. LLMs and diffusion models provide the commonsense knowledge and generalization, while NeRF is a tool for 3-D reconstruction and high fidelity scene rendering. Diffusion models also support the dynamic modeling, which is useful for world model building. It is seen that multi-modal language models can be built with a training dataset of sensor-text-action data (with the help of LLMs) to generate reasonable and realistic prediction for the world model and the simulator.

TABLE I
SIMULATORS AND WORLD MODELS

| Methods | Modalities | Functions | Technologies |
|---|---|---|---|
| READ[216] | image | Novel view synthesis | Neural rendering |
| Scene-Diffusion[219] | 2-D BEV | Traffic scene generation | Latent diffusion, object detection |
| MARS[223] | Image, depth, semantics | Photorealistic view synthesis | NeRF |
| UniSim[227] | Image, point cloud | Sensor simulation | NeRF, Hypernet |
| Adv3D[230] | image | Adversarial example generation | NeRF, 3D detector, Lift3D[203] |
| SurrealDrive[232] | Carla simulator, text | Driving behavior synthesis | LLM(GPT4), prompt, COT reasoning |
| DriveSceneGen[239] | 2-D BEV | 2-D BEV view synthesis | Diffus. model, Motion Transformer |
| MagicDrive[241] | Image, roadmap, object | Street view generation | Stable diffusion, ControlNet[140] |
| DrivingDiffusion[247] | Image, 2D layout | Multi-view videos generation | CLIP, diffusion model, prompt |
| RITA[212] | 2-D BEV | Traffic flow synthesis | Diffusion model, vehicle control |
| CTG[221] | 2-D BEV | Traffic flow synthesis | Diffusion model, STL rules |
| CTG++[222] | 2-D BEV | Traffic flow synthesis | Diffusion model, LLM (GPT4) |
| trafficBots[217] | 2-D BEV | World model for planning | RL, map, interaction, GRU |
| UniWorld[229] | LiDAR, multi-view images | World model for behavior perceiving and prediction | 4D occupancy, BEV encoder, finetuning, pre-trained |
| GAIA-1[240] | Video, text, actions | World model for driving scene generation, control over behavior | Sequence modeling, prediction, casual masking of attention, Transformer-based video decoder, diffusion model, text prompting |
| DriveDreamer[250] | 2-D BEV, text, action, image, HDMap, 3D box | World model for future video and action predictions | Diffusion model, CLIP, GRU, transformer |
| GPT-like WM[256] | Actions, point cloud | World model for GPT-like modeling and predicting future | MaskedGIT, VQVAE, diffus. model, BEV, depth rendering, tokenizer |
| MUVO[258] | Actions, images, point cloud | World model for forecasting the images, LiDAR point clouds & 3-D occupancy grids | Imitation learning, Transformer, sensor fusion, GRU |
| ADriver-I[259] | Actions, images | World model for predict the future frame and control signals | LLM, MLLM, video diffusion model, CLIP-ViT, prompt, fine-tuning |
| OccWorld[260] | point cloud/images | World model to predict the future frame and control signals | LLM, MLLM, video diffusion model, CLIP-ViT, prompt, fine-tuning |
| Drive-WM[261] | images | World model to predict images and apply for e2e planning | CNN, MLP, Diffusion model, VAE, cross attention, CLIP |

*B. Automatic Annotation (Perception Only)*

Data annotation is the cornerstone of deep learning model training, cause mostly model training runs in a supervised manner. Automatic labeling is strongly helpful for autonomous driving research and development, especially for open vocabulary scene annotations. LLMs and VLMs provide a way to realize it based on the learned knowledge and common sense.

Recently, models trained with large-scale image-text datasets have demonstrated robust flexibility and generalization capabilities for open-vocabulary image-based classification, detection and semantic segmentation tasks. Though it does not perform in real time, a human-level cognition capability is potential to behave like a teacher model at the cloud side, teaching a student model at client side to realize approximated performance.

Talk2Car[210] is an object referral dataset when taking into account the problem in an autonomous driving setting, where a passenger requests an action that can be associated with an object found in a street scene. It contains commands formulated in textual natural language for self-driving cars. The textual annotations are free form commands, which guide the path of an autonomous vehicle in the scene. Each command describes a change of direction, relevant to a referred object found in the scene. Similar works see CityScapes-Ref, Refer-KITTI and Nuscenes-QA etc.

OpenScene[213] is a simple yet effective zero-shot approach for open-vocabulary 3D scene understanding. The key idea is to compute dense features for 3D points that are co-embedded with text strings and image pixels in the CLIP feature space. To achieve this, associations between 3D points and pixels from posed images in the 3D scene are established, and a 3D network is trained to embed points using CLIP pixel features as supervision.

MSSG (Multi-modal Single Shot Grounding ) [220] is a multi-modal visual grounding method for LiDAR point cloud with a token fusion strategy, shown in Fig. 36. It jointly learns the LiDAR-based object detector with the language features and predicts the targeted region directly from the detector without any post-processing. The cross-modal learning enforces the detector to concentrate on important regions in the point cloud by considering the informative language expressions.

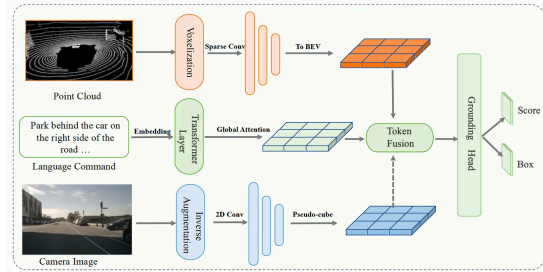

Fig. 36. MSSG [220]

HiLM-D (Towards High-Resolution Understanding in MLLMs for Autonomous Driving)[233] is an efficient method to incorporate HR information into MLLMs for the perception task. Especially, HiLM-D integrates two branches: (i) the LR reasoning branch, can be any MLLMs, processes LR videos to caption risk objects and discern ego-vehicle intentions/suggestions; (ii) the HR perception branch (HR-PB), ingests HR images to enhance detection by capturing vision-specific HR feature maps.

NuPrompt [235] is the object-centric language prompt set for driving scenes within 3D, multi-view, and multi-frame space. It expands Nuscenes dataset by constructing a total of 35,367 language descriptions, each referring to an average of 5.3 object tracks. Based on the object-text pairs from the new benchmark, a prompt-based driving task, i.e., employing a language prompt to predict the described object trajectory across views and frames is formulated. Furthermore, a simple end-to-end baseline model based on Transformer, named PromptTrack (modified from PF-Track, Past-and-Future reasoning for Tracking), is provided.

In [237] a multi-modal auto labeling pipeline is presented capable of generating amodal 3D bounding boxes and tracklets for training models on open-set categories without 3D human labels, defined as Unsupervised 3D Perception with 2D Vision-Language distillation (UP-VL), shown in Fig. 37. This pipeline exploits motion cues inherent in point cloud along with the freely available 2D image-text pairs. This method can handle both static and moving objects in the unsupervised manner and is able to output open-vocabulary semantic labels thanks to the proposed vision-language knowledge distillation.

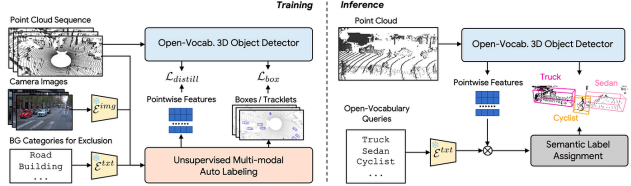

Fig. 37. UP-VL [237]

OpenAnnotate3D [252], is an opensource open-vocabulary auto-labeling system that can automatically generate 2D masks, 3D masks, and 3D bounding box annotations for vision and point cloud data, shown in Fig. 38. It integrates the chain-of-thought (CoT) capabilities of LLMs and the cross-modality capabilities of VLMs. Current off-the-shelf cross-modality vision-language models are based on 2D images, such as CLIP and SAM.

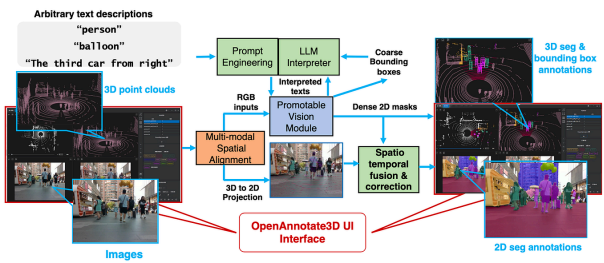

Fig. 38. OpenAnnotate3D [252]

Table II summarizes methods of automatic annotation. LLMs provide the knowledge to support labeling, meanwhile VLMs or MMLMs extend the LLMs to more modalities for annotating more diverse data.

TABLE II
AUTO ANNOTATION

| Methods | Modalities | Functions | Technologies |
| --- | --- | --- | --- |
| Talk2Car[210] | Text, action, object, image, point cloud, GPS+IMU | Object referral | Amazon Mechanical Turk (AMT), manual caption |
| OpenScene[213] | Point cloud, image, text | Open vocabulary 3D scene understanding | CLIP, segmentation, 2D-3D ensemble feature, zero-shot learning |
| MSSG[220] | LiDAR, text, image | Multi-modal visual ground. | Token fusing, object detector |
| HiLM-D[233] | Videos, text | Video understanding | LLMs, MMLMs, perception, reasoning, prompt, ViT, GradCAM |
| NuPrompt[235] | image, text | Language prompt generation | Transformer-based PromptTrack, LLM(GPT3.5), prompt, manual caption |
| UP-VL[237] | Image, point cloud | Auto labeling | Open set categories, VLMs, 3-D object detector, tracking |
| OpenAnnotate[252] | LiDAR, camera | Open vocabulary auto labeling | LLMs, VLMs, prompt, multi-modal spatial alignment |

### C. Decision making, Planning and E2E

#### 1. Large Scale Language Models' Integration

Drive-Like-a-Human [224] is a closed-loop system to showcase its abilities in driving scenarios (for instance, HighwayEnv, i.e. a collection of environments for *autonomous driving* and tactical decision-making tasks), by using an LLM (GPT3.5), shown in Fig. 39. Besides, perception tools and agent prompts are provided to aid its observation and decision-making. The agent prompts provide GPT-3.5 with information about its current actions, driving rules, and cautions. GPT-3.5 employs the ReAct strategy [82] to perceive and analyze its surrounding environment through a cycle of thought, action, and observation. Based on this information, GPT-3.5 makes decisions and controls vehicles in HighwayEnv, forming a closed-loop driving system.

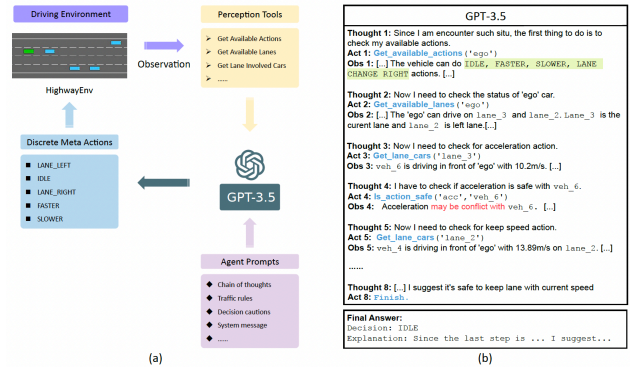

Fig. 39. Drive-Like-a-Human [224]

An open-loop driving commentator LINGO-1 [231] is proposed by Wayve which combines vision, language and action to enhance how to interpret and train the driving models, like Visual Question Answering (VQA). At Wayve, similar to embodied AI, vision-language-action models (VLAMs) are explored, which incorporate three kinds of information: images, driving data, and now language. The language can be used to explain the causal factors in the driving scene, which may enable faster training and generalization to new environments. VLAMs open up the possibility of interacting with driving models through dialogue, where users can ask autonomous vehicles what they are doing and why. In other words, LINGO-1 can provide a description of the driving actions and reasoning.

A text-based representation of traffic scenes is proposed [234] and processed with a pre-trained language encoder. Text-based representations, based on DistilBERT (a slim variant of BERT), combined with classical rasterized image representations, lead to descriptive scene embeddings, which are subsequently decoded into a trajectory prediction. Predictions on the nuScenes dataset is given as a benchmark.

Drive-as-You-Speak [236] applies LLMs to enhance autonomous vehicles' decision-making processes. It integrates LLMs' natural language capabilities and contextual understanding, specialized tools usage, synergizing reasoning, and acting with various modules on autonomous vehicles. It is aimed at seamlessly integrating the advanced language and reasoning capabilities of LLMs into autonomous vehicles.

A Reasoning module and a Reflection module is leveraged in DiLu[238] to run decision-making based on common-sense knowledge and improve continuously, shown in Fig. 40. To be specific, the Reasoning Module is used by the driver agent to query experiences from the Memory Module and the common-sense knowledge of the LLM is applied to build decisions based on current scenarios. Then the Reflection Module is applied to classify safe or unsafe decisions, and subsequently modify them into correct decisions by the knowledge embedded in the LLM.

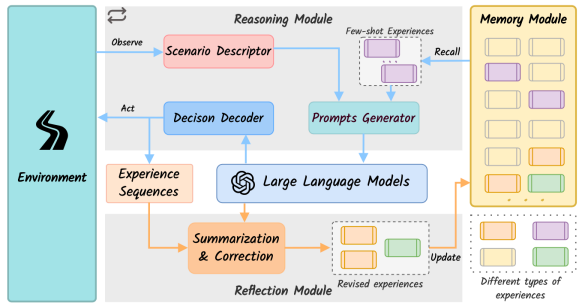

Fig. 40. DiLu [238]

LanguageMPC[242] employs LLMs as a decision-making component for complex AD scenarios that require human common-sense understanding. The cognitive pathways are designed to enable comprehensive reasoning with LLMs, as well as algorithms for translating LLM decisions into actionable driving commands. Through this approach, LLM decisions are seamlessly integrated with low-level controllers (MPC) by guided parameter matrix adaptation.

DriveGPT4[243] is an interpretable E2E autonomous driving system utilizing LLMs (LLaMA2). DriveGPT4 is capable of interpreting vehicle actions and providing corresponding reasoning, as well as answering diverse questions posed by human users for enhanced interaction. DriveGPT4 also predicts vehicle low-level control signals in an E2E fashion with the help of a customized visual instruction tuning dataset specifically designed for autonomous driving. Based on tokenization, the language model can concurrently generate responses to human inquiries and predict control signals for the next step. Upon producing predicted tokens, a de-tokenizer decodes them to restore human languages.

GPT-Driver[244] transforms the LLM (GPT3.5) into a reasonable motion planner of autonomous vehicles. It makes use of the strong reasoning capabilities and generalization potential in LLMs. The idea is to formulate motion planning as a language modeling problem, in which the planner inputs and outputs are converted into language tokens, and the driving trajectories are built through a language description of coordinate positions. Furthermore, it applies a prompting-reasoning-finetuning strategy to stimulate the numerical reasoning potential inherent in the LLM.

LLM-Driver[245] is an object-level multimodal LLM architecture that merges vectorized numeric modalities with a pre-trained LLM to improve context understanding in driving situations. A distinct pretraining strategy is devised to align numeric vector modalities with static LLM representations using vector captioning language data. As a matter of fact, training the LLM-Driver involves formulating it as a Driving Question Answering (DQA) problem within the context of a language model.

Talk2BEV[246] is a large vision language model (LVLM) interface for bird's-eye view (BEV) maps in autonomous driving contexts, based on BLIP-2[94] and LLaVA[87], shown in Fig. 41. The BEV map from image and LiDAR data is first generated. Then the language-enhanced map is constructed, augmented with aligned image-language features for each object from LVLMs. These features can directly be used as context to LVLMs for answering object-level and scene-level queries. Talk2BEV-Bench is a benchmark encompassing 1000 human annotated BEV scenarios, with more than 20,000 questions and ground-truth responses from the NuScenes dataset.

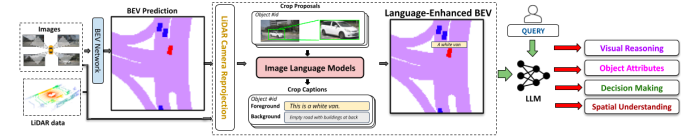

Fig. 41. Talk2BEV

DriveLM is an autonomous driving (**AD**) dataset incorporating linguistic information [248]. Through DriveLM, people can connect LLMs and autonomous driving systems, and eventually introduce the reasoning ability of LLM in AD to make decisions and ensure explainable planning. Specifically, in DriveLM, Perception, Prediction, and Planning (**P3**) are facilitated with human-written reasoning logic as a connection. To take it a step further, The idea of Graph-of-Thought (**GoT**) is leveraged to connect the QA pairs in a graph-style structure and use "What if"-style questions to reason about future events that have not happened.

Drive-Anywhere [254] is a generalizable E2E autonomous driving model with multimodal foundation models to enhance the robustness and adaptability, shown in Fig. 42. Specifically, it is capable of providing driving decisions from representations queried by image and text. To do so, a method is proposed to extract nuanced spatial (pixel/patch-aligned) features from Transformers (ViT) to enable the encapsulation of both spatial and semantic features. This approach allows the incorporation of latent space simulation (via text) for improved training (data augmentation via text by LLMs) and policy debugging.

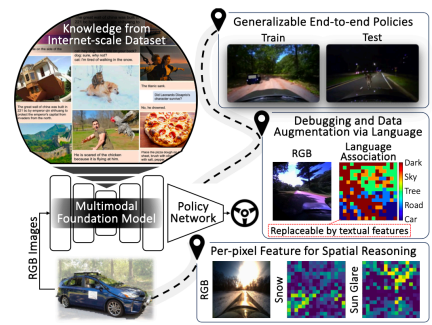

Fig. 42. Drive-Anywhere [254]

Agent-Driver[257], transforms the traditional autonomous driving pipeline by introducing a tool library accessible via function calls, a cognitive memory of common sense and experiential knowledge for decision-making, and a reasoning engine capable of chain-of-thought reasoning, task planning, motion planning, and self-reflection. Powered by LLMs, Agent-Driver is endowed with intuitive common sense and robust reasoning capabilities, thus enabling a more nuanced, human-like approach to autonomous driving.

Table III summarize the methods of LLM/VLM-based autonomous driving models. This is the mostly adapted way to integrate LLM/VLM into the self driving model, where the common sense of human knowledge is naturally applied for reasoning and decision making to handle driving policies and navigation.

TABLE III
LLM/VLM-BASED AUTONOMOUS DRIVING

| Methods | Modalities | Functions | Technologies |
|---|---|---|---|
| Drive-like-a-human[224] | 2-D BEV, text | Planning and control | LLM(GPT3.5), perception tool, ReAct[82], LLaMA-Adapter v2, prompts |
| LINGO-1[231] | Image, text, driv. data | Actions and reasoning like VQA | LLM, Vision Language Action Model |
| Can you text what is happening?[232] | Image, text | Trajectory prediction | LLM (DistilBERT) |
| Drive as you speak[236] | 2-D BEV, map, GNSS, radar, LiDAR, image | Decision making for actions | LLM (chatGPT4), special tool (localization, perception and monitoring) |
| DiLu[238] | text | Decision making to control | LLM(GPT3.5), prompt, recall from memory, CoT, decision decoder |
| languageMPC[242] | Text | Decision making for action commands | LLM(GPT3.5), CoT reasoning |
| DriveGPT4[243] | Control action, text, image, video | Action interpreting, e2e control, question answering | LLM (LLaMA2), tokenizer, de-tokenizer, visual instruction tuning |
| GPT-Driver[244] | Text, trajectory | Motion planner for trajectory generation and control | LLM(GPT3.5), prompt, fine tuning, tokenizer, perception & prediction tools, CoT reasoning |
| LLMDriver[245] | Text | Driving question answer | LLM (GPT3.5), pretrained model, RL expert, LoRA fine tuning |
| Talk2BEV[246] | BEV, image, LiDAR, text | Augment BEV map with language fir scene understanding, reasoning and decision making | LLM, Q-former in BLIP-2[87], LLaVA (instruction tuning)[94], question answering, BEV network |
| DriveLM[248] | 2-D BEV, text | Make decision and planning | LLM, perception, prediction & planning (P3), GoT, question answering |
| Drive-Anywhere[254] | Image, text | E2E driving policies with multi-modal understanding | LLM, BLIP[87], open set learning, attention mask, ViT[70], perception, policy net |
| Agent-Driver[257] | Image, text | E2E cognitive agent for autonomous driving | Tool library, cognitive memory, reasoning engine, self-reflection |

An obvious perspective to apply LLMs for autonomous driving is, LLMs can serve as the decision-making module, various functions, such as the perception module, localization module and prediction module, act as the vehicle's sensing device. Besides, the vehicle's actions and controller function as its executor, running orders from the LLM's decision-making process.

Similarly, a multi-modal language model is built from sensor-text-action data (with the help of LLMs) for end-to-end autonomous driving, either generate trajectory prediction or control signals directly, like a LLM instruction tuning solution. Another way to apply LLMs is merging vectorized modalities (encoded with input from raw sensor or tools like perception, localization and prediction) with a pre-trained LLM, like a LLM augmented solution.

*2. Tokenization like NLP's GPT*

A framework [211], called Talk-to the-Vehicle, consisting of a Natural Language Encoder (NLE), a Waypoint Generator Network (WGN) and a local planner, is designed to generate navigation waypoints for the self-driving car. NLE takes as input the natural language instructions and translates them into high-level machine-readable codes/encodings. WGN combines the local semantic structure with the language encodings to predict the local waypoints. The local planner generates an obstacle avoiding trajectory to reach the locally generated waypoints and executes it by employing a low-level controller.

ADAPT (Action-aware Driving cAPtion Transformer) [215], is an end-to-end transformer-based architecture, which provides user-friendly natural language narrations and reasoning for each decision making step of autonomous vehicular control and action. ADAPT jointly trains both the driving caption task and the vehicular control prediction task, through a shared video representation.

ConBaT(Control Barrier Transformer)[218] is an approach that learns safe behaviors from demonstrations in a self-supervised fashion, like a world model. ConBaT uses a causal transformer, derived from the Perception-Action Causal Transformer (PACT), that learns to predict safe robot actions autoregressively using a critic that requires minimal safety data labeling. During deployment, a lightweight online optimization is employed to find actions that ensure future states lie within the learned safe set.

The MTD-GPT (Multi-Task Decision-Making Generative Pre-trained Transformer) method [226] abstracts the multi-task decision making problem in autonomous driving as a sequence modeling task, shown in Fig. 43. Leveraging the inherent strengths of reinforcement learning (RL) and the sequence modeling capabilities of the GPT, it simultaneously manages multiple driving tasks, such as left turns, straight-ahead driving, and right turns at unsignalized intersections.

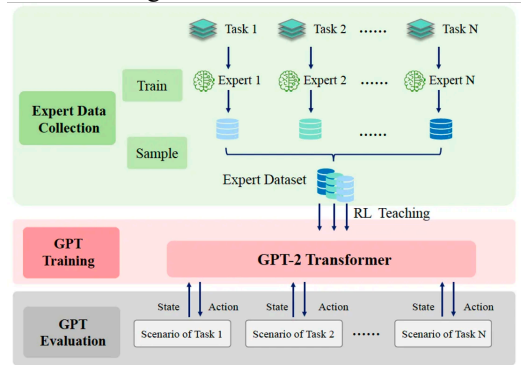

Fig. 43. MTD-GPT [226]

BEVGPT [251], shown in Fig. 44, is a generative pre-trained large model that integrates driving scenario prediction, decision-making, and motion planning. The model takes the bird's-eye-view (BEV) images as the only input source and makes driving decisions based on surrounding traffic scenarios. To ensure driving trajectory feasibility and smoothness, an optimization-based motion planning method is developed.

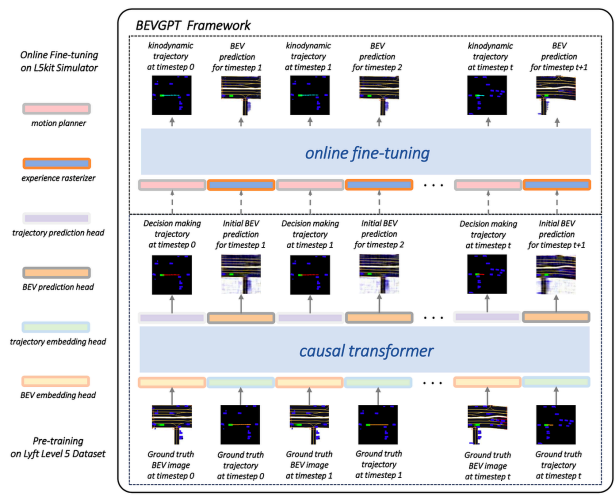

Fig. 44. BEVGPT [251]

Table IV summarizes the methods of tokenization like NLP's GPT. Instead of directly calling pretrained LLM/VLM, this type of approach builds the model based on self collected data (with the help of LLM/VLM) in a similar way as the language GPT.

TABLE IV
GPT-LIKE TOKENIZATION

| Methods | Modalities | Functions | Technologies |
|---|---|---|---|
| Talk-to-the-vehicle[211] | Depth, semantics, text | Navigation with trajectory generation to controller | CNN, LSTM, instruction, Natural Language Encoder, WGN, local planner |
| ADAPT[215] | videos | E2E decision making for control and action | Motion transformer, visual-language transformer for caption generation |
| ConBaT[218] | trajectory | Behavior (action) learning | Casual transformer, tokenizer, world model, control barrier |
| MTD-GPT[226] | Object trajectory (position and speed), action | Decision making | GPT-2Transformer-based RL, GPT-like sequence modeling, POMDP, policy network, action prediction |
| BEVGPT[251] | BEV images | Generative pretrained model with prediction, decision making and motion planning | Pre-training BEV prediction, modified GPT casual transformer, fine-tuning, optimization-based planning |

*3. Pre-trained Foundation Model*

PPGeo (Policy Pre-training via Geometric modeling)[214] is a fully self-supervised driving policy pre-training framework to learn from unlabeled and uncalibrated driving videos. It models the 3D geometric scene by jointly predicting ego-motion, depth, and camera intrinsics. In the first stage, the ego-motion is predicted based on consecutive frames as does in conventional depth estimation frameworks. In the second stage, the future ego-motion is estimated based on the single frame by a visual encoder, and could be optimized with the depth and camera intrinsics network well-learned in the first stage.

AD-PT (Autonomous Driving Pre-Training) [223] leverages the few-shot labeled and massive unlabeled point-cloud data to generate the unified backbone representations that can be directly applied to many baseline models and benchmarks, decoupling the AD-related pre-training process and downstream fine-tuning task. During this work a large-scale pre-training point-cloud dataset with diverse data distribution is built for learning generalizable representations.

UniPad[249] is a self-supervised learning paradigm applying 3D volumetric differentiable rendering, shown in Fig. 45. UniPAD implicitly encodes 3D space, facilitating the reconstruction of continuous 3D shape structures and the intricate appearance characteristics of their 2D projections. It can be seamlessly integrated into both 2D and 3D frameworks, enabling a more holistic comprehension of the scenes.

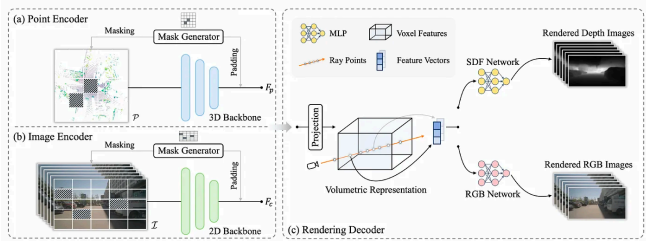

Fig. 45. UniPad [249]

Table V summarizes the methods of pretrained foundation models. This way seldom applies LLM/VLM info.

## VII. CONCLUSION

In simulation, we find combination of language model + diffusion model + NeRF will be the trend to realize photorealistic sensor data and human-like traffic flows. The similar thing happens to the world model, but it needs to model the environment behavior (especially the dynamics) due to the purpose of prediction.

TABLE V
PRETRAINED FOUNDATION MODELS

| Methods | Modalities | Functions | Technologies |
|---|---|---|---|
| PPGeo[214] | images | Policy pre-training | Posenet, depthnet, policy learning |
| AD-PT[223] | Point clouds | Unified pretrained representation learning | Mean Teacher, pseudo label generator, open set learning |
| UniPad[249] | Point clouds, images | Pretrained 3D representation learning | MAE-based mask generator, VoxelNet, differentiable neural rendering, SDF network |

In automatic annotation, multi-modal language models play important roles, especially for 3-D data. Mostly the visual-language model is the base, expanded to additional modalities with less data. The LLMs and VLMs provide the possibility of open vocabulary scene understandings.

In decision making and E2E, we still prefer the integration of large scale language models or multi-modal large scale language models. This category can be further split as three types, i.e. LLM as decision-making brain, LLM-augmented solution and LLM instruction tuning solution.

Either pretrained foundation models or tokenization like NLP's GPT sounds to be a strong owner of the autonomous driving large scale models, however the performance is more difficult to realize the grounding capabilities due to limited data collection and concern of hallucination.

The big issue is real time requirement in autonomous driving, so far no solutions with foundation models can afford this cost in hardware. Based on that, applications in autonomous driving will first appear at use cases as simulation and annotation.

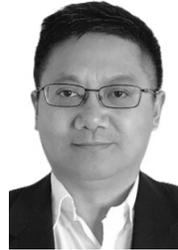

**Yu Huang**, CEO of Roboraction.AI. He was Chief Scientist of Synkrotron in 2022-2023, Chief AD Scientist in Zone Tech of SAIC Motor in 2022, Chief Scientist of Zhito Tech of FAW Motor in 2021-2022, VP of AD Research in Black Sesame (USA) in 2020-2021, also President of Singulato USA Inc. in 2018-2020. He has worked for Baidu USA R&D Center in 2016-2018, Intel in 2014-2016, Samsung USA R&D center in 2011-2012, Huawei USA R&D center in 2008-2011, and Thomson Corporate Research USA in 2005-2008 etc. He was postdoctoral associate at Beckman Inst. of UIUC in 2000-2002, and AvH research fellow at U. of Erlangen-Nuremberg in 1999-2000, postdoctoral fellow at Tsinghua U. in 1997-1999. He got Ph D at Beijing Jiaotong U. in 1997, MS degree at Xidian U. in 1993 and BS degree at Xi'an Jiaotong U. in 1990.

**Yu Chen**, has more than 20 years of experience in software development, product management, and technology strategy planning. He started his career in Santera Systems in 1999 and joined Huawei's American R&D center in 2009. From 2012 to 2019, he was the chief technology planning architect in Huawei's corporate technology planning office. From 2019 he is the VP of the technology planning and cooperation of Futurewei Technologies Inc. He received the BSEE from Beijing Institute of Technology and the MSEE from Chinese Academy of Sciences.

**Zhu Li** (Senior Member, IEEE), professor at U. of Missouri, Kansas city. He received his PhD in Electrical & Computer Engineering from Northwestern U., Evanston in 2004. He was Sr. Staff Researcher/Sr. Manager with Samsung Research America's Multimedia Standards Research Lab in Richardson, TX, 2012-2015, Sr. Staff Researcher/Media Analytics Group Lead with Huawei's Media Lab in Bridgewater, NJ, 2010~2012, and an Assistant Professor with the Dept of Computing, The Hong Kong Polytechnic U. in 2008-2010, and a Principal Staff Research Engineer with the Multimedia Research Lab (MRL), Motorola Labs, in 2000-2008. He has 46 issued or pending patents, 100+ academic publications.